\documentclass[preprint,12pt,3p]{elsarticle}

\usepackage{amssymb}
\usepackage{times}
\usepackage{epsfig}
\usepackage{graphicx}
\usepackage{amsmath}
\usepackage{subfigure}
\usepackage{algorithm}
\usepackage{algorithmic}
\usepackage{multirow}
\usepackage{color}
\usepackage{natbib}
\usepackage{caption}
\usepackage{slashbox}
\usepackage{booktabs}
\usepackage{multirow}
\usepackage{mathptmx}  %the below two lines are used for mathcal font correction
\usepackage{mathrsfs}
\DeclareMathAlphabet{\mathcal}{OMS}{cmsy}{m}{n}
\DeclareSymbolFont{largesymbols}{OMX}{cmex}{m}{n}
\usepackage{amsthm}
\usepackage{amsfonts}
\usepackage{url} %2015-8-5
\usepackage{textcomp}
\usepackage[pagebackref=true,breaklinks=true,letterpaper=true,colorlinks,bookmarks=false,citecolor=blue]{hyperref}
\journal{arXiv}

\begin{document}

\begin{frontmatter}

\title{Learning Deep Multi-Level Similarity for Thermal Infrared Object Tracking}
\author[label1]{Qiao Liu \corref{equ}}
\address[label1]{School of Computer Science and Technology, Harbin Institute of Technology Shenzhen Graduate School, China}
\author[label1]{Xin Li\corref{equ}}
\author[label1]{Zhenyu He\corref{cor1}}\ead{zhenyuhe@hit.edu.cn}
\author[label1]{Nana Fan}
\author[label1]{Di Yuan}
\author[label1]{Hongpeng Wang\corref{cor1}}\ead{wanghp@hit.edu.cn}

%\address[label2]{School of Computer Science and Technology, Harbin Institute of Technology, Harbin, China}
\cortext[cor1]{I am corresponding author}
\cortext[equ]{Contribution equally}

\begin{abstract}
Existing deep Thermal InfraRed (TIR) trackers only use semantic features to describe the TIR object, which lack the sufficient discriminative capacity for handling distractors. This becomes worse when the feature extraction network is only trained on RGB images.
To address this issue, we propose a multi-level similarity model under a Siamese framework for robust TIR object tracking. Specifically, we compute different pattern similarities on two convolutional layers using the proposed multi-level similarity network. One of them focuses on the global semantic similarity and the other computes the local structural similarity of the TIR object. These two similarities complement each other and hence enhance the discriminative capacity of the network for handling distractors. In addition, we design a simple while effective relative entropy based ensemble subnetwork to integrate the semantic and structural similarities. This subnetwork can adaptive learn the weights of the semantic and structural similarities at the training stage. To further enhance the discriminative capacity of the tracker, we construct the first large scale TIR video sequence dataset for training the proposed model. The proposed TIR dataset not only benefits the training for TIR tracking but also can be applied to numerous TIR vision tasks. Extensive experimental results on the VOT-TIR2015 and VOT-TIR2017 benchmarks demonstrate that the proposed algorithm performs favorably against the state-of-the-art methods.
\end{abstract}


\end{frontmatter}

% \linenumbers

\section{Introduction}
Thermal InfraRed (TIR) object tracking is an important branch of visual object tracking, which receives more and more attention recently. Compared with visual tracking~\cite{zhang2019parallel,PatchTracking,BMtracker,zhang2018visual,Multi-viewCF,WLStracker,MOT,group-lasso,peng2016hybrid,BIAtracker}, TIR tracking has several superiorities such as the illumination insensitivity and privacy protection.
Since the TIR tracking method can track the object in total darkness, it has a wide range of applications such as video surveillance, maritime rescue, and driver assistance at night~\cite{TC}. However, there are several problems faced in TIR tracking are still challenging such as occlusion, appearance changed, and similar distractors.

%review some related works
To handle various challenges, numerous TIR trackers are proposed by researchers in the past decade. For instances, TBOOST~\cite{TBOOST} ensembles several MOSSE filters~\cite{MOSSE} using a continuously switching mechanism to choose a set right base trackers. TBOOST can adapt the appearance variation of the object, since it maintains a dynamics ensemble.
Sparse-tir~\cite{l1-tir-tracker} explores the sparse representation with a compressive Harr-like features for real-time TIR tracking, which can handle the occlusion challenge to some extent due to the feature of spare representation. Similar to Sparse-tir, MF-tir~\cite{multi-feature} also uses the sparse representation method for TIR tracking but explores multiple complemental features for getting more discriminative feature representation.
DSLT~\cite{DSLT} uses an online structural support vector machine~\cite{struck} with a combination of the motion feature and a modified Histogram of Oriented Gradient (HOG)~\cite{HOG} feature for TIR tracking. DSLT obtains the favorable performance mainly because the dense online learning and more robust feature representation.
There are also a series of TIR trackers are proposed based on  kernel density estimation~\cite{liu2012infrared}, multiple instances learning~\cite{shi2013infrared}, low-rank sparse learning~\cite{he2016infrared}, discriminative correlation filter~\cite{he2015infrared,asha2017robust}, etc.
Despite these methods achieve much progress, they performances are limited by the hand-crafted feature representation.

%background and problem
Recently, inspired by the success of Convolution Neural Network (CNN) in visual tracking~\cite{FCNT,zhang2016robust,HCF,C-COT}, several methods explore CNN for TIR tracking. DSST-tir~\cite{16KTIR} shows that deep features are more effective than  hand-crafted features for TIR tracking. MCFTS~\cite{MCFTS} uses a pre-trained VGGNet~\cite{VGGNet} to extract deep feature and combine Correlation Filter (CF)~\cite{KCF} to achieve an ensemble TIR tracker.
LMSCO~\cite{LMSCO} integrates CF and structural support vector machine using a combination of the deep appearance feature and the deep motion feature for TIR tracking.
HSSNet~\cite{HSSNet} trains a verification based Siamese CNN on RGB images for TIR tracking. However, most of these methods just use a deep semantic feature. Unlike the visual object, the TIR object does not have color information or rich texture features, which makes it difficult to distinguish TIR objects belonging to a same class. This shows that only using a global semantic feature is insufficient for handling distractors in TIR tracking. Furthermore, most of these deep TIR trackers are trained on RGB images due to lacking a large scale TIR image training dataset, which further degrades the discriminative capacity.

%method
To address the above-mentioned problems, we propose a multi-level similarity model, called MLSSNet, under a Siamese framework for robust TIR tracking. We note that the multi-level similarity is effective in enhancing the discriminative capacity of the Siamese network for handling distractors.  To this end, we design a structural Correlation Similarity Network (CSN) and a semantic CSN to compute different pattern similarities on different convolutional layers.
The structural CSN captures the local structural information of a pair of TIR objects and then computes the structural similarity of them. We identify that the structure information can help the network distinguish TIR objects belonging to a same class.
The semantic CSN enhances the global semantic representation capacity and then computes the similarity on the semantic level.
%The similarity of the bottom layers that contains the local structural information, and the similarity of the higher layers that contains the global semantic information.
To obtain an optimal comprehensive similarity containing the structural and semantic similarities simultaneously, we design a Relative Entropy based adaptive ensemble Network (REN) to integrate them.
Furthermore, to enhance the discriminative capacity of MLSSNet, we construct a large scale TIR image training dataset with manual annotations. The dataset has $430$ videos with a total of over $180,000$ TIR frames and over $200,000$ bounding boxes. We note that the tracker has a more powerful discriminative capacity for handling distractors when it is trained on the TIR images dataset.
We analyze the multi-level similarity model with an ablation study and compare it with state-of-the-art methods on the VOT-TIR2015~\cite{VOT-TIR2015} and VOT-TIR2017~\cite{VOTTIR2017} benchmarks in Section~\ref{ablation} and Section~\ref{comparison} respectively. The favorable performance against the state-of-the-arts demonstrates the effectiveness of the proposed method.

%contribution
The contributions of the paper are three-fold:
\begin{itemize}
  \item We propose a multi-level similarity model under the Siamese network framework for robust TIR object tracking. The network consists of three specialized designed subnetworks which are the structural CSN, the semantic CSN, and the REN.
  \item We construct the first large scale TIR image training dataset
  %~\footnote{If the paper is accepted, we will release the dataset.}
   with manual annotations. The dataset can be used to train the deep network for solving several TIR vision tasks.
  \item We carry out extensive experiments on benchmarks and demonstrate that the proposed TIR tracker performs favorably against state-of-the-art methods.
\end{itemize}

%rest framework.
The rest of the paper is organized as follows. We first introduce related tracking methods and TIR training dataset briefly in Section~\ref{RW}. Then, we describe the architecture and training details of the proposed multi-level similarity network in Section~\ref{MLSS}. Subsequently, the extensive experiments are reported in Section~\ref{exp} to show the proposed method achieves favorable performance. Finally, we draw a short conclusion and describe some future work in Section~\ref{conclusion}.

\section{Related Work}
\label{RW}
In this section, we first introduce the Siamese framework based trackers, which are most related with ours. Then, we discuss the several ensemble learning strategies in CNN tracking method. Finally, we describe several TIR training datasets used for tracking.
\vspace{2mm}

\noindent{\textbf{Siamese based trackers.}}
%Object tracking problem is treated as a similarity verification task that is popular in the community recently. Most common practice is to off-line train a similarity metric network and then uses it to online compute the similarity between candidates and the target.
Siamese based trackers treat object tracking as a similarity verification task, most of which is to off-line train a similarity metric network and then uses it to online compute the similarity between candidates and the target.
%Siamese architecture and its variants~\cite{MSiam} are often used to achieve the above-mentioned task in object tracking. The first simple yet effective fully convolutional Siamese network: Siamese-FC~\cite{Siamese-fc} is proposed since 2016, several variations with better performance are quickly presented to handle the various challenges.
For example, Siamese-FC~\cite{Siamese-fc} trains the first fully convolutional Siamese network for tracking and achieves promising results.
In order to adapt the appearance variation of the target, DSiam~\cite{DSiam} learns a dynamic Siamese network by two line regression models. One of these models can learn the target's appearance change and the other can learn to suppress the background. StructSiam~\cite{StructSiam} learns a structured Siamese network, which focuses on the local pattern of the target and their structural relationship.
To obtain more powerful features, SiamFC-tri~\cite{SiamFC-tri} uses a triplet loss to train the Siamese network, which learns the triplet relationship instead of the pairwise relationship. SA-Siam~\cite{SA-Siam} exploits a twofold Siamese network which is composed by a semantic branch and an appearance branch. These two branches are trained from different tasks to complement each other. Our method also uses two branches that is similar to SA-Siam~\cite{SA-Siam} but there are several significant differences. First, SA-Siam uses two separate branches trained with different tasks to compute different similarity, while our model uses two branches trained with one task (the same loss function) to compute different similarity. Second, the two branches of SA-Siam are trained separately, while ours is trained end-to-end. Third, the two branches are fused in the tracking stage via a simple add operation in SA-Siam, while ours are fused in the training stage using a relative entropy-based adaptive ensemble network.
In order to enhance the discriminative capacity of the Siamese network, CFNet~\cite{CFNet} introduces CF as a differentiable layer into the Siamese network. This layer can update the target branch using video-specific cues that could be helpful for discrimination. RASNet~\cite{RASNet} introduces residual attention into CFNet to further boost the discriminative capacity. Considering the motion information is helpful for tracking, FlowTrack~\cite{FlowTrack} trains an optical flow network and a CFNet model simultaneously.
To achieve high performance and high speed simultaneously, SiamRPN~\cite{SiamRPN} employs a Siamese region proposal network which consists of a feature extraction subnetwork and a region proposal subnetwork. It is formulated as a local one-shot detection task in the tracking stage. Subsequently, DaSiamRPN~\cite{DaSiamRPN} extends SiamRPN by controlling the distribution of the training data and achieves top performance in the VOT2018~\cite{VOT2018} challenge.
However, most of these methods compute similarity from one single level e.g., the semantic level. Different from these methods, in this paper, we exploit the multi-level similarity to enhance the discriminative capacity of the Siamese network for handling distractors in TIR tracking.

\vspace{2mm}
\noindent{\textbf{CNN based ensemble trackers.}}
%CNN based ensemble trackers receive much attention recently.
The ensemble learning is used at different stages in object tracking.  For example, HDT~\cite{HDT} combines the multiple weak CNN based CF trackers into a stronger one by a Hedge algorithm. This algorithm can adaptively update the weights of each weak tracker. STCT~\cite{STCT} trains an ensemble based CNN classifier for tracking via a sequential sampling method. Similar to STCT, Branchout~\cite{Branchout} trains an ensemble based CNN classifier by using a stochastic regularization technology. TCNN~\cite{TCNN} manages the multiple CNNs in a tree structure to estimate target states and to update the model.
EDCF~\cite{EDCF} integrates a low-level fine-grained feature and a high-level semantic feature in a mutually reinforced way.
Though both the proposed REN and the HDT methods use an adaptive ensemble strategy, the proposed REN model is trained end-to-end, which fuses multiple similarities at the learning stage.

\vspace{2mm}
\noindent{\textbf{TIR training dataset.}} The lacking of a large scale TIR image training dataset hinders the development of CNN in TIR object tracking. Several methods attempt to train a CNN model on TIR dataset for TIR tracking. For instance, DSST-tir~\cite{16KTIR} investigates the deep CNN feature in CF for TIR object tracking. This CNN model is trained on a small scale TIR image dataset with the classification task. It experimental results show that the deep feature based CF tracker can obtain better performance than hand-crafted feature based CF tracker. ECO-tir~\cite{SyntheticTIR} trains a Generative Adversarial Network~\cite{pix2pix} (GAN) to generate synthetic TIR images from RGB images. These synthetic images, the number of which is over 80,000, are used to train CFNet~\cite{CFNet} end to end for feature extraction of the TIR object. It experimental results show that more TIR training data contributes to better performance. In this paper, we construct a large scale real TIR image sequences from references and video websites including GRB-T~\cite{RGB-T}, PTB-TIR~\cite{PTB-TIR}, OSU~\cite{OSU}, OTCBVS~\cite{OTCBVS}, PDT-ATV~\cite{PDT-ATV}, and YouTube~\cite{Youtube} with  manual annotations for training the proposed model. To the best of our knowledge, we are the first to construct a large scale real TIR training dataset.

\begin{figure*}[t]
\begin{center}
\includegraphics[width=1\textwidth]{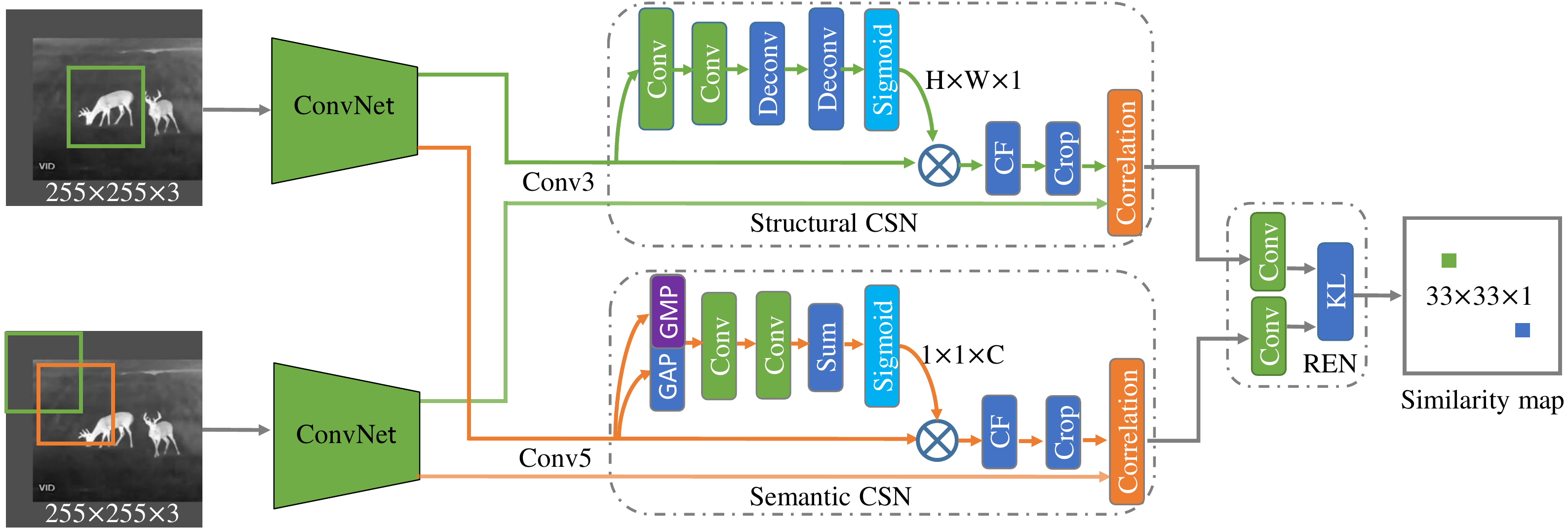}
\end{center}
\caption{Architecture of the proposed Multi-Level Similarity based Siamese Network (MLSSNet). MLSSNet is constituted by a shared feature extractor, a structural Correlation Similarity Network (CSN), a semantic CSN, and an adaptive fusion model (REN). Every block denotes a specific network layer and each convolution layer joins a hidden ReLU layer. GAP, GMP, CF, KL, and $\bigotimes$ denote the global average pooling, global max pooling, correlation filter, Kullback-Leibler divergence , and scale layer respectively. }
\label{architecture}
\end{figure*}

\section{Multi-Level Similarity Network}
\label{MLSS}
In this section, we first describe the framework of the proposed multi-level similarity network in Section~\ref{framework}, which is mainly consists of three specific designed subnetworks: structural CSN, semantic CSN, and REN. Then, we introduce the training details of the network in  Section~\ref{training}, which including the TIR training dataset and loss function. Finally, we present how to use the proposed network for TIR tracking in Section~\ref{trackinginterface}

\subsection{Network architecture}
\label{framework}
To achieve more effective TIR tracking, we construct a multi-level similarity model under Siamese framework, as shown in Figure~\ref{architecture}. Unlike existing Siamese network which often computes the similarity based on one feature space, we compute the similarity from multiple levels including the local structure level and the global semantic level. We note that the multi-level similarity can improve the discriminative capacity of the Siamese network, and hence improve the robustness of the TIR tracker. To this end, we design two different subnetworks: structural CSN and semantic CSN to compute the local structural similarity and the global semantic similarity respectively. Furthermore, we design a simple while effective adaptive ensemble subnetwork: REN to integrate the structural similarity and semantic similarity. In the following, we highlight these specific networks in details.

\begin{figure*}[t]
\centering
\subfigure{\label{feature-soccer}\includegraphics [width=1\textwidth]{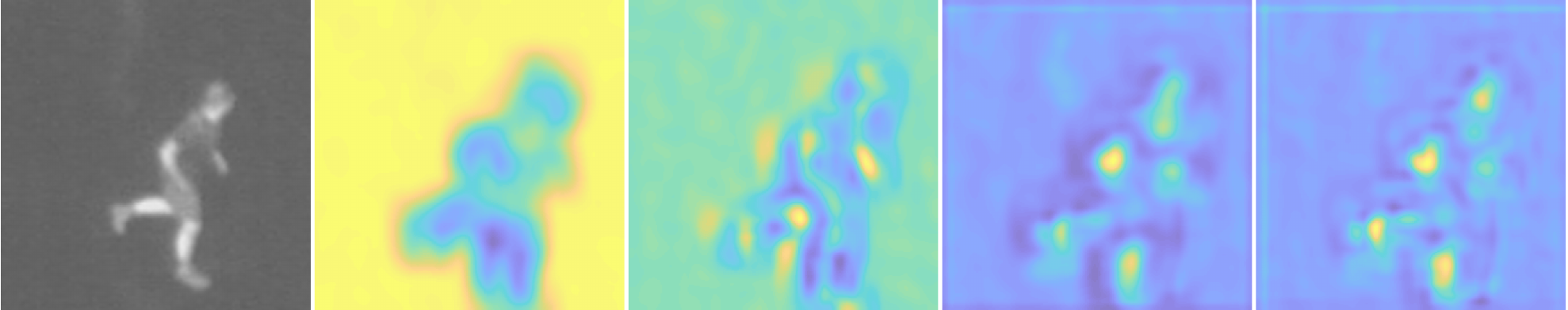}}\\
\vspace{-0.1in}
\begin{subfigure}{\label{feature-rhino}\includegraphics [width=1\textwidth]{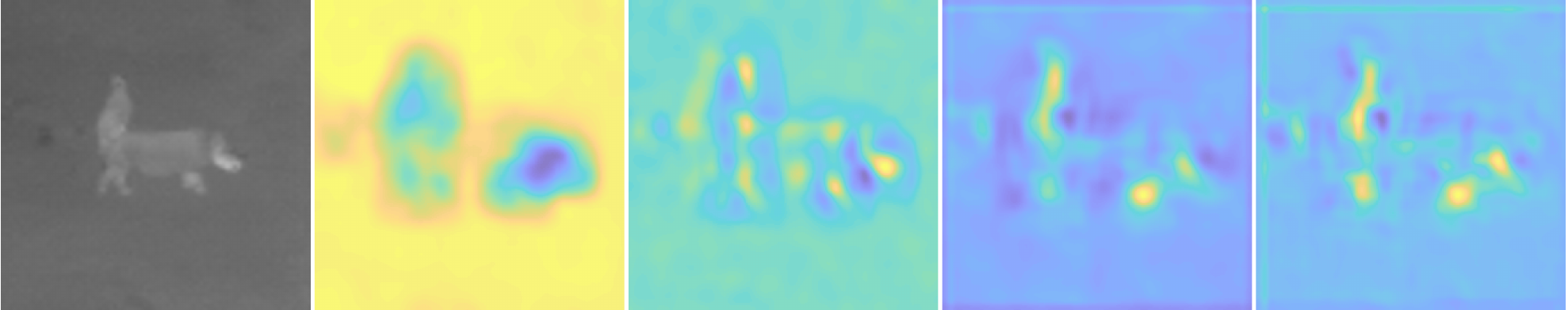}}
%\subcaption*{Input images \qquad \qquad \quad Conv3  \qquad \qquad Structure-aware \qquad \qquad Conv5  \qquad  \qquad Semantic-aware }
\end{subfigure}
\caption{Visualization of the original and the learned structure-aware and semantic-aware features. The visualized feature maps are generated by summing all channels. From left to right on each column are the input images, the original feature on Conv3, the learned structure-aware feature form Conv3, the original feature on Conv5, and the learned semantic-aware feature from Conv5 respectively. We can see that the structure-aware features tend to focus on the local structure parts, e.g., head and leg, while the semantic-aware features emphasize the more discriminative global semantic regions. }
\label{featuremap}
\end{figure*}

\vspace{2mm}
\noindent{\textbf{Structural CSN.}} To compute the structural similarity, we design a structure-aware subnetwork to capture the local structure feature of the object on the shallow convolution layer. We note that the local structure similarity is helpful for the accurate location of the tracker. Since the TIR objects lack the color and texture information, the local structure feature is crucial for the tracker to distinguish them. Specifically, we first use two convolution layers with kernel sizes of $7\times7$ and $5\times5$ to capture the local structure information of the object on the shallow layer of CNN. Then, we locate these structure parts by using two deconvolution layers with kernel sizes of $5\times5$ and $7\times7$, respectively. Next, we use a Sigmoid layer to generate a two dimension weight map which indicates the importance of every local structure. Finally, we use a scale layer to weight the original feature via the weight map. The weighted feature is aware of the local structure of the object, as shown in Figure~\ref{featuremap}. After the scale layer, we add a CF~\cite{CFNet} layer to update the target template. Given an input image $\mathbf{z}$ and a search image $\mathbf{x}$ , the structural similarity can be formulated as:
 \begin{equation}\label{structureCSN}
   f_{struct}(\mathbf{z},\mathbf{x})=Corr(\varphi(\omega(\phi_{conv3}(\mathbf{z}))),\phi_{conv3}(\mathbf{x})),
 \end{equation}
where $\phi_{conv3}(\cdot)$ denotes the third convolutional features of the shared feature extraction network, $\omega(\cdot)$ represents the structure-aware subnetwork, $\varphi(\cdot)$ is the CF block~\cite{CFNet}, and $Corr(\cdot,\cdot)$ denotes the cross-correlation operator.

\vspace{2mm}
\noindent{\textbf{Semantic CSN.}} To compute the semantic similarity, we design a semantic-aware subnetwork to enhance the semantic representation ability of the deep convolution feature. Since the discriminative capacity of the network mainly comes from semantic feature, it is important to obtain a more powerful semantic feature. To this end, our semantic-aware subnetwork formulates the relationship of feature channels to generate more powerful feature, which is similar to SENet~\cite{SENet}. Specifically, we first squeeze the feature map into two one dimension vectors by a global average pooling and a global max pooling respectively. Then, we use two shared convolution layers to formulate the relationship between these channels and then we fuse the two kinds of relationship vectors via a Sum layer. Different from previous methods, we use two kinds of global pooling because we note that they provide different clues for the global semantic information.
Next, we use a Sigmoid layer to generate a one dimension weight vector which indicates the importance of each feature channel. Finally, we employ a scale layer to weight the origin feature via the weight vector. The weighed feature emphasizes the discriminative region and hence obtains more powerful semantic feature representation, as shown in Figure~\ref{featuremap}. Similar to the structural similarity, the semantic similarity can be formulated as:
\begin{equation}\label{semantic}
   f_{semantic}(\mathbf{z},\mathbf{x})=Corr(\varphi(\nu(\phi_{conv5}(\mathbf{z}))),\phi_{conv5}(\mathbf{x})),
\end{equation}
where $\phi_{conv5}(\cdot)$ denotes the fifth convolutional features of the shared feature extraction network, $\nu(\cdot)$ represents the semantic-aware subnetwork.

\vspace{2mm}
\noindent{\textbf{REN.}} To integrate the structural similarity and semantic similarity, we propose an adaptive ensemble subnetwork which is constituted by two $1 \times 1$ convolution layers and a specific designed Kullback-Leibler (KL) divergence layer. The aim of this subnetwork is to obtain a comprehensive similarity map which has a minimum distance from the structural and semantic similarities. Given $n$ similarity maps $S=\{S^{1},S^{2},\cdots,S^{n}\}$, we hope to get an integrated similarity map $Q\in \mathbb{R}^{M\times N}$. Since the each similarity map can be regarded as a probability distribution of the object target, we can use KL divergence to measure the distance between the similarity map $S^{k}(k=1,2,\cdots,n)$ and the integrated similarity map $Q$. Then, we minimize the distance to optimize the similarity map $Q$ by:
\begin{equation}\label{KLMIN}
  \mathop{\arg\min}_{Q}\sum_{k=1}^{n}{KL(S^{k}\|Q)}
  \quad s.t.\sum{q_{ij}=1,}
\end{equation}
where
\begin{equation}\label{KL}
  KL(S^{k}\|Q)=\sum_{ij}s_{ij}^{k}\log \frac{s_{ij}^{k}}{q_{ij}},
\end{equation}
$s_{ij}$ and $q_{ij}$ denote the $(i,j)th$ element of the similarity map $S$ and $Q$ respectively. We use the Lagrange multiplier method to solve Eq.~\ref{KLMIN} and the solution has a simple formulation as:
\begin{equation}\label{solution}
  Q=\frac{1}{n}\sum_{k=1}^{n}{S^{k}}.
\end{equation}
Therefore, the KL layer can be regraded as a weighted sum operator. According to Eq.~\ref{solution}, the final integrated similarity can be formulated by:
\begin{equation}\label{ensemble}
f(\mathbf{z},\mathbf{x})=\frac{1}{2}( \alpha f_{struct}(\mathbf{z},\mathbf{x}) + \beta f_{semantic}(\mathbf{z},\mathbf{x}))+ b,
\end{equation}
where $\alpha$ and $\beta$ denote the parameter of the two convolution filters respectively. $b$ is the sum of bias of the two convolution layers. These parameters are learned adaptively.

\begin{figure*}[t]
\begin{center}
\includegraphics[width=1\textwidth]{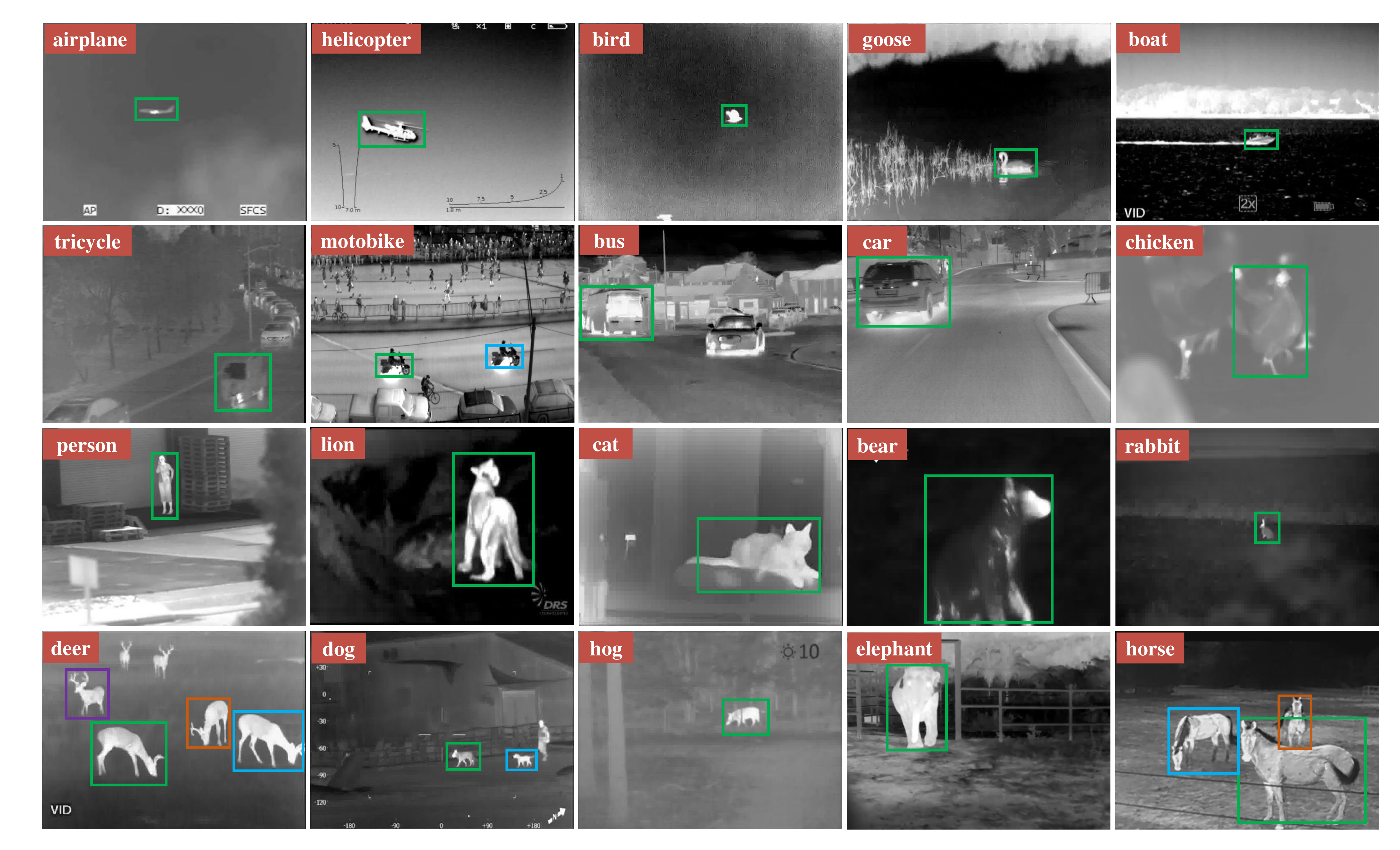}
\end{center}
\caption{Examples of our TIR video training dataset. We annotate the class name and bounding box location of the object in each frame of the video. Some videos have multiple objects such as the deer, horse, and dog sequences.}
\label{TIRdataset}
\end{figure*}

\subsection{Network training}
\label{training}
\noindent{\textbf{TIR training dataset.}} To further enhance the performance of the proposed method, we construct a TIR video training dataset, as shown in Figure~\ref{TIRdataset}, for training the proposed network. The dataset contains $430$ TIR videos with a total of over $180,000$ images. We annotate the location and class name of the object in each frame according to the ILSVRC2015~\cite{ILSVRC} format. The dataset has $20$ classes and over $200,000$ bounding boxes.
Since the videos are collected from references and video websites, the source and resolution of the images are various. Furthermore, the shot scene and shot time of the videos are also various. Therefore, the dataset has real data distribution and high diversity. Table~\ref{attribute} shows several attributes of the proposed TIR video dataset. All of the images are shown as the white-hot mode and stored with a 8 bits depth.  Most of videos are shotted at night, thus, the most object targets are warmer than its background.
From Table~\ref{attribute}, we can see that the proposed TIR dataset is captured from four kinds of devices, such as hand-held camera, surveillance (static) camera, vehicle-mounted camera, and drone camera. Therefore, the dataset contains most real-world challenges including occlusion, size change, camera motion, motion blur, and dynamic change.
\begin{table}[htbp]
\centering
\caption{Some attributes of the constructed TIR video training dataset. }
\begin{tabular}{l|r}
\toprule
Attribute & Value    \\
\midrule
Video sequences  & 430   \\
Class numbers     & 20     \\
Total Images     & Over 180,000 \\
Total bounding boxes & Over 200,000 \\
Image depth           & 8 bits    \\
Image resolution      & $320\times 240$ to $1280\times 720$ \\
Shot categories     & Surveillance, Hand-held, Vehicle-mounted, Drone \\
\bottomrule
\end{tabular}
\label{attribute}
\end{table}

\vspace{2mm}
\noindent{\textbf{Training samples generation.}} As shown in Figure~\ref{architecture}, the network needs a pair of cropped samples as inputs. First, we mix the ILSVRC2015 dataset with our TIR dataset.
Then, we convert RGB images of ILSVRC2015 to grayscale since the TIR object does not have color information.
Finally, we crop the image and choose the positive and negative training pairs from the whole training dataset like in CFNet~\cite{CFNet}.

\vspace{2mm}
\noindent{\textbf{Loss function.}} We use the logistic loss to train the proposed network. Since the similarity map measures the similarity between a target and multiple candidates, the loss function should be a mean loss:
\begin{equation}\label{meanloss}
  L(y,o)=\frac{1}{\mid \mathcal{D} \mid}\sum_{u \in \mathcal{D}}\log(1+exp(-y[u]o[u])),
\end{equation}
where $\mathcal{D} \in \mathbb{R}^{2}$ denotes the similarity map, $o[u]$ represents the real score of a single target-candidate pair and $y[u]$ is the ground-truth of this pair.

\subsection{Tracking interface}
\label{trackinginterface}
After training of the proposed model, we just use it as a match function at the tracking stage without any online updating. Given a target image $\mathbf{z}_{t-1}$ at the $(t-1)$-th frame and a search region $\mathbf{x}_{t}$ at the $t$-th frame, the tracked target at the $t$-th frame can be formulated by:
\begin{equation}\label{tracking}
  \hat{\mathbf{x}}_{t,i}= \mathop{\arg\max}_{\mathbf{x}_{t,i}} f(\mathbf{z}_{t-1},\mathbf{x}_{t}),
\end{equation}
where $\mathbf{x}_{t,i} \in \mathbf{x}_{t}$ is the $i$-th candidate in the search region $\mathbf{x}_{t}$. To handle the scale variation of the object, we use a simple scale estimation strategy like that in~\cite{Siamese-fc}.

\section{Experiments}
\label{exp}
In this section, we first present the implementation details in Section~\ref{details}. Then, we analyse the effectiveness of each component of the proposed method in Section~\ref{ablation}. Finally, we compare our approach with state-of-the-art methods in Section~\ref{comparison}.

\subsection{Experimental details}
\label{details}
We use a modified AlexNet~\cite{AlexNet} as the shared feature extractor. We add paddings $(1, 1)$ on the last two convolution layers since we need to output two aligned similarity maps by the structural CSN and semantic CSN. Before using the semantic CSN, we reduce channel number of the third convolution layer to $64$. We train MLSSNet via Stochastic Gradient Descent (SGD) with the momentum of $0.9$ and weight decay of $0.0005$ using MatConvNet~\cite{matconvnet}. The learning rate exponentially decays from $10^{-2}$ to $10^{-5}$. The network is trained for $50$ epochs and we set the mini-batch size to $8$. In the tracking stage, we set three fixed scales to $\{0.9745, 1, 1.0375\}$ for handling scale variation of the object. The current scale is updated by a linear interpolation with a factor of $0.59$ on the predicted scale. The proposed method is carried out on a PC with a GTX 1080 GPU card and achieves a speed of $18$ frames per second.

\subsection{Ablation studies}
\label{ablation}
To demonstrate that each component of the proposed method is effective, we compare our method with its variants on the VOT-TIR2015~\cite{VOT-TIR2015} and VOT-TIR2017~\cite{VOTTIR2017} benchmarks.

\vspace{2mm}
\noindent{\textbf{Datasets.}} VOT-TIR2015~\cite{VOT-TIR2015} is a first standard TIR tracking benchmark which provides the dataset and toolkit to fair evaluate TIR trackers. The dataset contains $20$ TIR image sequences and six kinds of challenges such as Dynamics Change (DC), Occlusion (Occ), Camera Motion (CM), Motion Change (MC), Size Change (SC), and Empty (Emp). VOT-TIR2017~\cite{VOTTIR2017} has $25$ TIR image sequences, which is more challenging than VOT-TIR2015. It also has six kinds of challenges which can be used to evaluate the corresponding performance of a tracker.

\vspace{2mm}
\noindent{\textbf{Evaluation criteria.}} Accuracy (Acc) and Robustness (Rob)~\cite{ARplot} are often used to evaluate the performance of a tracker from different aspects due to the high interpretability~\cite{VOTperformance1}. While accuracy is computed from the overlap rate between the prediction and ground truth, robustness is measured in term of the frequency of tracking failure. Furthermore, there is a comprehensive evaluation criterion called Expected Average Overlap (EAO)~\cite{VOT2015} which is adopted to measure the overall performance of a tracker.

\begin{table}[htbp]
\centering
\caption{Comparison of MLSSNet and its variants on VOT-TIR2015 and VOT-TIR2017. The up arrow and down arrow denote the bigger or smaller value is, the better corresponding performance has. }
\begin{tabular}{l|rrr|rrr}
\toprule
& \multicolumn{3}{c|}{VOT-TIR2015} & \multicolumn{3}{c}{VOT-TIR2017} \\
\midrule
Tracker & EAO $\uparrow$ & Acc $\uparrow$ & Rob $\downarrow$  & EAO $\uparrow$ & Acc $\uparrow$ & Rob $\downarrow$  \\
\midrule
         Baseline &0.282 & 0.55  & 2.82 & 0.254 & 0.52 &3.45 \\
         Baseline\_Sem & 0.299 & 0.54 & 2.49 &0.258  &0.53 &3.27 \\
         Baseline\_Sem+Str & 0.309 & 0.56 &2.60  &0.272 &0.54 &3.36 \\
         MLSSNet & \textbf{0.316} &\textbf{0.57} & \textbf{2.32}  & \textbf{0.278} & \textbf{0.56} & \textbf{2.95} \\
\bottomrule
\end{tabular}
\label{ablationT}
\end{table}

\vspace{2mm}
\noindent{\textbf{Results and analysis.}} We use CFNet trained on ILSVRC2015 as the baseline method. First, to show that the semantic CSN is effective, we compare Baseline with its variation (Baseline\_Sem) adding the semantic CSN module. The results of Table~\ref{ablationT} show that the semantic CSN improves the robustness of the baseline method on both two benchmarks remarkably. This demonstrates that the semantic CSN can enhance the discriminative capacity of the original feature representation. Second, to demonstrate that the structural CSN is effective, we compare Baseline\_Sem with Baseline\_Sem+Str which denotes that the structural CSN is added into Baseline\_Sem. The results of Table~\ref{ablationT} show that the structural CSN enhances the accuracy and EAO of Baseline\_Sem about two and one percent on VOT-TIR2015 respectively. This shows that the local structural similarity is helpful for precisely object location.  Third, to show that the proposed TIR dataset can further enhance the discriminative capacity, we compare Baseline\_Sem+Str and MLSSNet which represents Baseline\_Sem+Str is trained on ILSVRC2015 and our TIR dataset simultaneously. The results of Table~\ref{ablationT} show that training on the TIR dataset boosts the accuracy and robustness of Baseline\_Sem+Str on two benchmarks remarkably.

\begin{table}[t]
\centering
\caption{Comparison of MLSSNet and state-of-the-arts on VOT-TIR2017 and VOT-TIR2015. The up arrow and down arrow denote the bigger or smaller value is, the better corresponding performance has. The bold and underline denote the best and the second-best respectively.}
\begin{tabular}{l|rrr|rrr}
\toprule
 & \multicolumn{3}{c|}{VOT-TIR2017} & \multicolumn{3}{c}{VOT-TIR2015} \\
\midrule
  Tracker & EAO $\uparrow$ & Acc $\uparrow$ & Rob $\downarrow$  & EAO $\uparrow$ & Acc $\uparrow$ & Rob $\downarrow$\\
\midrule
  MCFTS~\cite{MCFTS}  & 0.193 & 0.55 & 4.72  &0.218 &0.59 & 4.12 \\
  HDT~\cite{HDT}& 0.196 & 0.51 & 4.93  &0.188 &0.53 &5.22   \\
  deepMKCF~\cite{DeepMKCF}  & 0.213 & 0.61 & 3.90 &- &- &-  \\
  Siamese-FC~\cite{Siamese-fc}  & 0.225 & 0.57 & 4.29 &0.219 &0.60 &4.10   \\
  SiamRPN~\cite{SiamRPN}                       & 0.242 &0.60  &3.19  &0.267 &0.63 &2.53\\
  MDNet-N~\cite{VOT-TIR2016}  & 0.243 & 0.57 & 3.33   &-  &- &- \\
  CREST~\cite{CREST}  & 0.252 & 0.59 & 3.26 &0.258 &0.62 &3.11  \\
  CFNet~\cite{CFNet}   & 0.254 & 0.52 & 3.45 &0.282 &0.55 &2.82  \\
  DaSiamRPN~\cite{DaSiamRPN}                     & 0.258 &0.62  &2.90  &0.311 &0.67 &2.33\\
  HSSNet~\cite{HSSNet}   & 0.262 & 0.58 &3.33 &\underline{0.311} &\textbf{0.67} &2.53 \\
  DeepSTRCF~\cite{DeepSTRCF}             &0.262  &0.62  &3.32 &0.257 &0.63 &2.93\\
  Staple-TIR~\cite{VOT-TIR2016}& 0.264 & \textbf{0.65} & 3.31   &- &- &-  \\
  ECO-deep~\cite{ECO}                      &0.267  &0.61 &\underline{2.73}  &0.286 &\underline{0.64} &2.36 \\
  MOSSE\_CA~\cite{VOTTIR2017}                     &0.271  &0.56 & - & - &- &-\\
  VITAL~\cite{VITAL}                          &0.272  &\underline{0.64} &\textbf{2.68} &0.289 &0.63 &\textbf{2.18} \\
  MLSSNet-No-TIR-D (Ours)  &  \underline{0.272} &  0.54 &  3.36  &0.309 &0.56 &2.60\\
  MLSSNet (Ours)   & \textbf{0.278} & 0.56  & 2.95 &\textbf{0.316} &0.57 &\underline{2.32} \\
\bottomrule
\end{tabular}
\label{SOTA}
\end{table}

\subsection{Comparison with state-of-the-arts}
\label{comparison}
To evaluate the proposed algorithm comprehensively, we compare our method with fifteen state-of-the-art methods on the VOT-TIR2017~\cite{VOTTIR2017} and VOT-TIR2015~\cite{VOT-TIR2015} benchmarks.

\vspace{2mm}
\noindent{\textbf{Dataset and evaluation criteria.}}  We also use the VOT-TIR2017 and VOT-TIR2015 benchmarks as the testing datasets and the Accuracy (Acc), Robustness (Rob), and EAO as the evaluation criteria.

\vspace{2mm}
\noindent{\textbf{Compared trackers.}} We compare the proposed methods (MLSSNet and MLSSNet-No-TIR-D which is trained without the proposed TIR dataset) with fifteen trackers. These methods can be divided into four categories. Six trackers are based on the deep correlation filter such as deepMKCF~\cite{DeepMKCF},  HDT~\cite{HDT}, MCFTS~\cite{MCFTS}, CREST~\cite{CREST}, ECO-deep~\cite{ECO}, and DeepSTRCF~\cite{DeepSTRCF}. Five trackers are based on the Siamese network framework such as Siamese-FC~\cite{Siamese-fc}, CFNet~\cite{CFNet}, SiamRPN~\cite{SiamRPN}, DaSiamRPN~\cite{DaSiamRPN}, and HSSNet~\cite{HSSNet}. Two hand-crafted feature based CF tracker: Staple-TIR~\cite{VOT-TIR2016}, MOSSE\_CA~\cite{VOTTIR2017}. Two deep trackers using classification and adversarial learning respectively:  MDNet-N~\cite{VOT-TIR2016} and VITAL~\cite{VITAL}.

\begin{table*}[t]
\centering
\caption{Comparison of MLSSNet and state-of-the-arts on the six challenges of the VOT-TIR2017 benchmark under EAO evaluation criterion.  DC, MC, CM, SC, Occ, Emp represent the dynamics change, motion change, camera motion, size change, occlusion, and empty respectively. The bold and underline denote the best and the second-best respectively.}
\begin{tabular}{l|rrrrrr}
\toprule
\backslashbox{Tracker\kern+6em}{Challenge}  & DC & MC & CM &SC &Occ &Emp  \\
\midrule
  MCFTS~\cite{MCFTS}      & 0.072 & 0.278 & 0.156 & 0.212 & 0.181 & 0.033   \\
  HDT~\cite{HDT}        & 0.090 & 0.258 & 0.167 & 0.243 & 0.181 & 0.034   \\
  deepMKCF~\cite{DeepMKCF}   & 0.074 & 0.319 & 0.179 & 0.255 & 0.189 & 0.035   \\
  Siamese-FC~\cite{Siamese-fc} & 0.188 & 0.319 & 0.196 & 0.277 & 0.222 & 0.043   \\
  SiamRPN~\cite{SiamRPN}       &0.167 &0.339 &0.215 &0.303 &0.226 &0.045 \\
  MDNet-N~\cite{VOT-TIR2016}    & 0.209 & 0.381 & 0.216 & 0.290 & 0.280 & 0.047   \\
  CREST~\cite{CREST}      & 0.130 & 0.348 & \textbf{0.256} & 0.300 & 0.278 & 0.036   \\
  CFNet~\cite{CFNet}      & \underline{0.222} & 0.410 & 0.219 & 0.285 & \underline{0.306} & \underline{0.058}   \\
  DaSiamRPN~\cite{DaSiamRPN} &0.114 &0.336 &0.208 &0.309 &0.214 &0.053 \\
  HSSNet~\cite{HSSNet}     & 0.204 & 0.430 & 0.204 & 0.309 & \textbf{0.317} & 0.050   \\
  DeepSTRCF~\cite{DeepSTRCF} &0.217 &0.359 &0.233 &\textbf{0.370} &0.268 &0.046 \\
  Staple-TIR~\cite{VOT-TIR2016} & 0.164 & 0.414 & 0.186 & 0.342 & 0.258 & 0.056   \\
  ECO-deep~\cite{ECO}  &0.192 &0.387 &0.233 &\underline{0.344} &0.280 &0.055 \\
  VITAL~\cite{VITAL}   &0.157 &\textbf{0.440} &\underline{0.254} &0.299 &0.253 &0.050 \\
  MLSSNet-No-TIR-D (Ours) &0.220 &0.436 &0.232 &0.320 &0.288 &0.053 \\
  MLSSNet (Ours)    & \textbf{0.251} & \underline{0.438} & 0.230 & 0.329 & 0.297 & \textbf{0.063}   \\
\bottomrule
\end{tabular}
\label{AttributeEAO}
\end{table*}

\vspace{2mm}
\noindent{\textbf{Results and analysis.}} First, we compare the comprehensive performance of our method with other trackers, as shown in Table~\ref{SOTA}. It shows that our approach achieves the best EAO of $0.278$ and $0.316$ on the VOT-TIR2017 and VOT-TIR2015 benchmarks respectively. We also see that although the proposed method does not use the constructed TIR training dataset, it also achieves the best and the second-best EAO $0.272$ and $0.309$ on the VOT-TIR2017 and VOT-TIR2015 benchmarks respectively. These results demonstrate that the proposed method performs favorably against these state-of-the-art methods.
Compared with the Siamese framework based tracker CFNet~\cite{CFNet}, our approach improves the accuracy and robustness simultaneously. The enhancement of performance is mostly attributed to the proposed multi-level similarity network.
Compared with CF based deep tracker CREST~\cite{CREST}, our method has better robustness despite CREST online updates the target template. We argue that the superior robustness of our method comes from the training on the constructed TIR dataset.
Though MDNet-N~\cite{VOT-TIR2016} online trains a deep classification network for TIR tracking, our method gets better robustness, which benefits from both the multi-level similarity and the constructed TIR training dataset.

\begin{table*}[t]
\centering
\caption{Comparison of MLSSNet and state-of-the-arts on the six challenges of the VOT-TIR2015 benchmark under EAO evaluation criterion.  DC, MC, CM, SC, Occ, Emp represent the dynamics change, motion change, camera motion, size change, occlusion, and empty respectively. The bold and underline denote the best and the second-best respectively.}
\begin{tabular}{l|rrrrrr}
\toprule
\backslashbox{Tracker\kern+6em}{Challenge}  & DC & MC & CM &SC &Occ &Emp  \\
\midrule
  MCFTS~\cite{MCFTS}      & 0.707 & 0.257 & 0.362 & 0.214 & 0.483 & 0.024   \\
  HDT~\cite{HDT}        & 0.689 & 0.224 & 0.215 & 0.187 & 0.463 & 0.027   \\
  Siamese-FC~\cite{Siamese-fc} & 0.671 & 0.319 & 0.226 & 0.273 & 0.406 & 0.033   \\
  SiamRPN~\cite{SiamRPN}       &\underline{0.725} &0.316 &0.372 &0.352 &0.404 &0.035 \\
  CREST~\cite{CREST}      & 0.708 & 0.331 & \underline{0.475} & 0.278 & \textbf{0.652} & 0.028   \\
  CFNet~\cite{CFNet}      & 0.590 & 0.387 & 0.459 & 0.326 & 0.495 & \textbf{0.050}   \\
  DaSiamRPN~\cite{DaSiamRPN} &0.718 &0.369 &0.370 & \underline{0.443} &0.457 &0.043 \\
  HSSNet~\cite{HSSNet}     & 0.661 & 0.426 & 0.407 & 0.383 & 0.496 & \underline{0.045}   \\
  DeepSTRCF~\cite{DeepSTRCF} &0.693 &0.337 &0.277 &0.354 &0.492 &0.035 \\
  ECO-deep~\cite{ECO}  &\textbf{0.745} &0.371 &0.335 &0.417 &0.614 &0.043 \\
  VITAL~\cite{VITAL}   &0.435 &0.392 &\textbf{0.561} &0.358 &0.526 &0.034 \\
  MLSSNet-No-TIR-D (Ours) &0.699 &\underline{0.444} &0.436 &0.382 &0.603 &0.044 \\
  MLSSNet (Ours)    & 0.694 & \textbf{0.480} & 0.429 & \textbf{0.453} & \underline{0.637} & 0.042   \\
\bottomrule
\end{tabular}
\label{AttributeEAO15}
\end{table*}

Second, in order to show the effectiveness of the proposed method for handling different challenging, we compare the proposed method with state-of-the-art methods on the six challenging attributes on the VOT-TIR2017 and VOT-TIR2015 benchmarks, as shown in Table~\ref{AttributeEAO} and Table~\ref{AttributeEAO15} respectively. It is obviously shown that our method achieves the best EAO on two challenges of VOT-TIR2017, which are the dynamics change and the empty. When the proposed method is trained on the constructed TIR training dataset, it's EAO gains about $3\%$ on the dynamics change challenge. This demonstrate that the TIR training dataset is important for enhancing the robustness of the tracker.
The proposed method also obtains the best and the second-best performance on the motion change of VOT-TIR2015 and VOT-TIR2017 respectively. Compare with CFNet~\cite{CFNet}, our two methods enhance EAO about $5\%$ and $3\%$ on the motion change of VOT-TIR2015 and VOT-TIR2017 respectively. This shows that the proposed multi-level similarity model is helpful for precise localization on the motion change challenge.
It is easy to see that our method achieves the competitive performance on the size change challenge of two benchmarks. Compare with Siamese-FC~\cite{Siamese-fc}, our tracker improves EAO about $ 18\%$ and $5\%$ on size change of VOT-TIR2015 and VOT-TIR2017 respectively. This demonstrates that the multi-level similarity model improves the robustness of the Siamese network remarkably, since these two trackers use a same scale estimation strategy.
These attributes based results demonstrate that our approach achieves a powerful discriminative capacity and favorable robustness.

\begin{figure}[htbp]
\centering
\subfigure{\label{big-bbs}\includegraphics [width=1\textwidth]{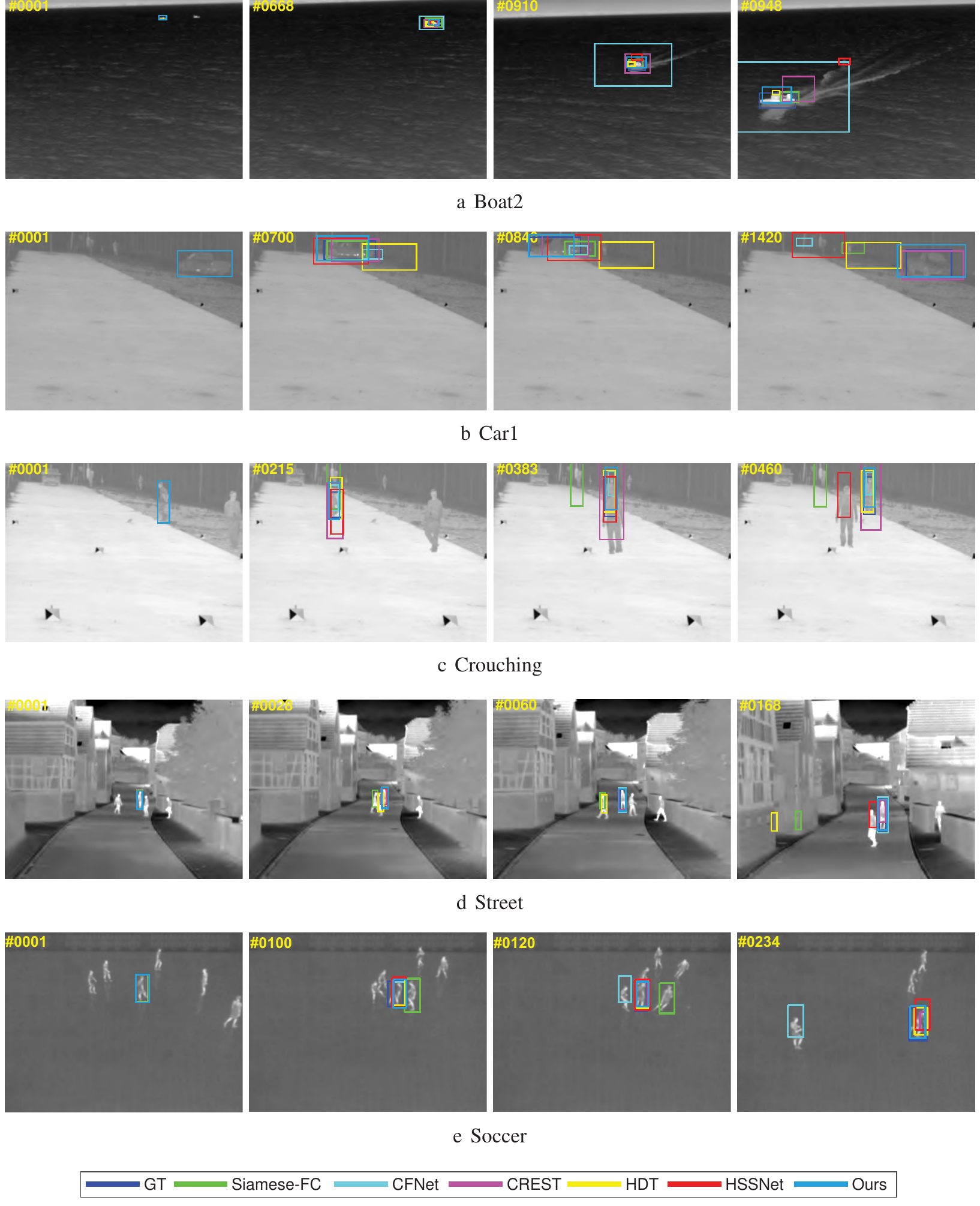}}
\caption{Tracking results visualized comparison of our tracker and state-of-the-arts on several challenging sequences.}
\label{bbs}
\end{figure}

Third, to visualized present the tracking performance, we show the tracking results of several trackers on five challenging sequences, as shown in Figure~\ref{bbs}. When the boat in "Boat2" changes the scale drastically, our tracker can more precisely locate the object than the others. Most of the methods fail when the car changes its appearance severely in "Car1", our method tracks it more accurately. This shows that our method has a more powerful discriminative capacity than others. When the Siamese framework based trackers Siamese-FC and HSSNet drift away from the target in the "Street" and "Soccer", our approach distinguishes the object from several distractors. This also shows that the multi-level similarity can improve the discriminative capacity of the Siamese network for handling distractors. While the appearance variation and distractor occurred in the "Crouching" simultaneously, the proposed method also achieves more accurate localization.

\section{Conclusion}
\label{conclusion}
This paper proposes a multi-level similarity model under the Siamese framework for robust Thermal InfraRed (TIR) object tracking. The network consists of a multi-level similarity network and a relative entropy based adaptive ensemble network.  The structural correlation similarity network captures the local structure information of the TIR object for precise location. While the semantic correlation similarity network enhances the global semantic representation of the feature for robust identification. The multi-level similarity improves the discriminative capacity of the Siamese network. In addition, to further enhance the discriminative capacity, we construct a large scale TIR image dataset to train the proposed model. The dataset not only benefits the training for TIR tracking but also can be applied to numerous TIR vision tasks such as classification and detection.
Extensive experimental results on the VOT-TIR2015 and VOT-TIR2017 benchmarks show that the proposed method performs favorably against the state-of-the-art methods.

\section*{Acknowledgment}
This research was supported by the National Natural Science Foundation of China (Grant No.61672183), by the Shenzhen Research Council (Grant Nos. JCYJ20170413104556946, JCYJ20\\160406161948211,
JCYJ20160226201453085), by the Natural Science Foundation of Guangdong Province (Grant No. 2015A030313544).

%% References
%% Usage of ~\cite is as follows:
%%   ~\cite{key}         ==>>  [#]
%%   ~\cite[chap. 2]{key} ==>> [#, chap. 2]
%%


%\bibliographystyle{elsarticle-num}
%% \bibliographystyle{elsarticle-harv}
%% \bibliographystyle{elsarticle-num-names}
%% \bibliographystyle{model1a-num-names}
%% \bibliographystyle{model1b-num-names}
%% \bibliographystyle{model1c-num-names}
%% \bibliographystyle{model1-num-names}
%% \bibliographystyle{model2-names}
%% \bibliographystyle{model3a-num-names}
%% \bibliographystyle{model3-num-names}
%% \bibliographystyle{model4-names}
%% \bibliographystyle{model5-names}
%% \bibliographystyle{model6-num-names}
%
%\bibliography{tracking}

\begin{thebibliography}{10}
\expandafter\ifx\csname url\endcsname\relax
  \def\url#1{\texttt{#1}}\fi
\expandafter\ifx\csname urlprefix\endcsname\relax\def\urlprefix{URL }\fi
\expandafter\ifx\csname href\endcsname\relax
  \def\href#1#2{#2} \def\path#1{#1}\fi

\bibitem{zhang2019parallel}
K.~Zhang, J.~Fan, Q.~Liu, J.~Yang, W.~Lian, Parallel attentive correlation
  tracking, IEEE Transactions on Image Processing 28~(1) (2019) 479--491.

\bibitem{PatchTracking}
Z.~He, S.~Yi, Y.-M. Cheung, X.~You, Y.~Y. Tang, Robust object tracking via key
  patch sparse representation, IEEE transactions on cybernetics 47~(2) (2017)
  354--364.

\bibitem{BMtracker}
S.~Zhang, X.~Lan, Y.~Qi, P.~C. Yuen, Robust visual tracking via basis matching,
  IEEE Transactions on Circuits and Systems for Video Technology 27~(3) (2016)
  421--430.

\bibitem{zhang2018visual}
K.~Zhang, X.~Li, H.~Song, Q.~Liu, W.~Lian, Visual tracking using
  spatio-temporally nonlocally regularized correlation filter, Pattern
  Recognition 83 (2018) 185--195.

\bibitem{Multi-viewCF}
X.~Li, Q.~Liu, Z.~He, H.~Wang, C.~Zhang, W.-S. Chen, A multi-view model for
  visual tracking via correlation filters, Knowledge-Based Systems 113 (2016)
  88--99.

\bibitem{WLStracker}
S.~Zhang, H.~Zhou, F.~Jiang, X.~Li, Robust visual tracking using structurally
  random projection and weighted least squares, IEEE Transactions on Circuits
  and Systems for Video Technology 25~(11) (2015) 1749--1760.

\bibitem{MOT}
Z.~He, X.~Li, X.~You, D.~Tao, Y.~Y. Tang, Connected component model for
  multi-object tracking, IEEE Transactions on Image Processing 25~(8) (2016)
  3698--3711.

\bibitem{group-lasso}
X.~Ma, Q.~Liu, W.~Ou, Q.~Zhou, Visual object tracking via coefficients
  constrained exclusive group lasso, Machine Vision and Applications 29~(5)
  (2018) 749--763.

\bibitem{peng2016hybrid}
Q.~Peng, Y.-m. Cheung, X.~You, Y.~Y. Tang, A hybrid of local and global
  saliencies for detecting image salient region and appearance, IEEE
  Transactions on Systems, Man, and Cybernetics: Systems 47~(1) (2016) 86--97.

\bibitem{BIAtracker}
S.~Zhang, X.~Lan, H.~Yao, H.~Zhou, D.~Tao, X.~Li, A biologically inspired
  appearance model for robust visual tracking, IEEE transactions on neural
  networks and learning systems 28~(10) (2016) 2357--2370.

\bibitem{TC}
R.~Gade, T.~B. Moeslund, Thermal cameras and applications: a survey, Machine
  vision and applications 25~(1) (2014) 245--262.

\bibitem{TBOOST}
E.~Gundogdu, H.~Ozkan, H.~S. Demir, H.~Ergezer, E.~Akagunduz, S.~K. Pakin,
  Comparison of infrared and visible imagery for object tracking: Toward
  trackers with superior ir performance, in: IEEE Conference on Computer Vision
  and Pattern Recognition (CVPR) Workshop, 2015, pp. 1--9.

\bibitem{MOSSE}
D.~S. Bolme, J.~R. Beveridge, B.~A. Draper, Y.~M. Lui, Visual object tracking
  using adaptive correlation filters, in: IEEE Computer Society Conference on
  Computer Vision and Pattern Recognition, 2010, pp. 2544--2550.

\bibitem{l1-tir-tracker}
Y.~Li, P.~Li, Q.~Shen, Real-time infrared target tracking based on $\ell_{1}$
  minimization and compressive features, Applied optics 53~(28) (2014)
  6518--6526.

\bibitem{multi-feature}
S.~J. Gao, S.~T. Jhang, Infrared target tracking using multi-feature joint
  sparse representation, in: International Conference on Research in Adaptive
  and Convergent Systems, 2016, pp. 40--45.

\bibitem{DSLT}
X.~Yu, Q.~Yu, Y.~Shang, H.~Zhang, Dense structural learning for infrared object
  tracking at 200+ frames per second, Pattern Recognition Letters 100 (2017)
  152--159.

\bibitem{struck}
S.~Hare, S.~Golodetz, A.~Saffari, V.~Vineet, M.-M. Cheng, S.~L. Hicks, P.~H.
  Torr, Struck: Structured output tracking with kernels, IEEE transactions on
  pattern analysis and machine intelligence 38~(10) (2016) 2096--2109.

\bibitem{HOG}
N.~Dalal, B.~Triggs, Histograms of oriented gradients for human detection, in:
  IEEE Computer Society Conference on Computer Vision and Pattern Recognition,
  Vol.~1, 2005, pp. 886--893.

\bibitem{liu2012infrared}
R.~Liu, Y.~Lu, Infrared target tracking in multiple feature pseudo-color image
  with kernel density estimation, Infrared Physics \& Technology 55~(6) (2012)
  505--512.

\bibitem{shi2013infrared}
X.~Shi, W.~Hu, Y.~Cheng, G.~Chen, J.~J.~H. Ling, Infrared target tracking using
  multiple instance learning with adaptive motion prediction and spatially
  template weighting, in: Proc. of SPIE Vol, Vol. 8739, 2013, pp. 873912--1.

\bibitem{he2016infrared}
Y.~He, M.~Li, J.~Zhang, J.~Yao, Infrared target tracking based on robust
  low-rank sparse learning, IEEE Geoscience and Remote Sensing Letters 13~(2)
  (2016) 232--236.

\bibitem{he2015infrared}
Y.-J. He, M.~Li, J.~Zhang, J.-P. Yao, Infrared target tracking via weighted
  correlation filter, Infrared Physics \& Technology 73 (2015) 103--114.

\bibitem{asha2017robust}
C.~Asha, A.~Narasimhadhan, Robust infrared target tracking using discriminative
  and generative approaches, Infrared Physics \& Technology 85 (2017) 114--127.

\bibitem{FCNT}
L.~Wang, W.~Ouyang, X.~Wang, H.~Lu, Visual tracking with fully convolutional
  networks, in: IEEE International Conference on Computer Vision (ICCV), 2015,
  pp. 3119--3127.

\bibitem{zhang2016robust}
K.~Zhang, Q.~Liu, Y.~Wu, M.-H. Yang, Robust visual tracking via convolutional
  networks without training, IEEE Transactions on Image Processing 25~(4)
  (2016) 1779--1792.

\bibitem{HCF}
C.~Ma, J.-B. Huang, X.~Yang, M.-H. Yang, Hierarchical convolutional features
  for visual tracking, in: IEEE International Conference on Computer Vision
  (ICCV), 2015, pp. 3074--3082.

\bibitem{C-COT}
M.~Danelljan, A.~Robinson, F.~S. Khan, M.~Felsberg, Beyond correlation filters:
  Learning continuous convolution operators for visual tracking, in: European
  Conference on Computer Vision, 2016, pp. 472--488.

\bibitem{16KTIR}
E.~Gundogdu, A.~Koc, B.~Solmaz, et~al., Evaluation of feature channels for
  correlation-filter-based visual object tracking in infrared spectrum, in:
  IEEE Conference on Computer Vision and Pattern Recognition Workshops (CVPRW),
  2016, pp. 24--32.

\bibitem{MCFTS}
Q.~Liu, X.~Lu, Z.~He, et~al., Deep convolutional neural networks for thermal
  infrared object tracking, Knowledge-Based Systems 134 (2017) 189--198.

\bibitem{VGGNet}
K.~Simonyan, A.~Zisserman, Very deep convolutional networks for large-scale
  image recognition, arXiv preprint arXiv:1409.1556.

\bibitem{KCF}
J.~F. Henriques, R.~Caseiro, P.~Martins, J.~Batista, High-speed tracking with
  kernelized correlation filters, IEEE Transactions on Pattern Analysis and
  Machine Intelligence 37~(3) (2015) 583--596.

\bibitem{LMSCO}
P.~Gao, Y.~Ma, K.~Song, C.~Li, F.~Wang, L.~Xiao, Large margin structured
  convolution operator for thermal infrared object tracking, in: IEEE
  International Conference on Pattern Recognition (ICPR), 2018, pp. 2380--2385.

\bibitem{HSSNet}
X.~Li, Q.~Liu, N.~Fan, et~al., Hierarchical spatial-aware siamese network for
  thermal infrared object tracking, Knowledge-Based Systems 166 (2019) 71--81.

\bibitem{VOT-TIR2015}
M.~Felsberg, A.~Berg, et~al., The thermal infrared visual object tracking
  vot-tir2015 challenge results, in: IEEE International Conference on Computer
  Vision (ICCV) Workshops, 2015, pp. 76--88.

\bibitem{VOTTIR2017}
M.~Kristan, A.~Leonardis, J.~Matas, M.~Felsberg, et~al., The visual object
  tracking vot2017 challenge results, in: IEEE International Conference on
  Computer Vision Workshops (ICCVW), 2017, pp. 1949--1972.

\bibitem{Siamese-fc}
L.~Bertinetto, J.~Valmadre, et~al., Fully-convolutional siamese networks for
  object tracking, in: European Conference on Computer Vision Workshops
  (ECCVW), 2016, pp. 850--865.

\bibitem{DSiam}
Q.~Guo, W.~Feng, C.~Zhou, et~al., Learning dynamic siamese network for visual
  object tracking, in: IEEE International Conference on Computer Vision (ICCV),
  2017, pp. 1763--1771.

\bibitem{StructSiam}
Y.~Zhang, L.~Wang, J.~Qi, et~al., Structured siamese network for real-time
  visual tracking, in: European Conference on Computer Vision (ECCV), 2018, pp.
  351--366.

\bibitem{SiamFC-tri}
X.~Dong, J.~Shen, Triplet loss in siamese network for object tracking, in:
  European Conference on Computer Vision (ECCV), 2018, pp. 459--474.

\bibitem{SA-Siam}
A.~He, C.~Luo, X.~Tian, W.~Zeng, A twofold siamese network for real-time object
  tracking, in: IEEE Conference on Computer Vision and Pattern Recognition
  (CVPR), 2018, pp. 4834--4843.

\bibitem{CFNet}
J.~Valmadre, L.~Bertinetto, et~al., End-to-end representation learning for
  correlation filter based tracking, in: IEEE Conference on Computer Vision and
  Pattern Recognition (CVPR), 2017, pp. 5000--5008.

\bibitem{RASNet}
Q.~Wang, Z.~Teng, et~al., Learning attentions: residual attentional siamese
  network for high performance online visual tracking, in: IEEE Conference on
  Computer Vision and Pattern Recognition (CVPR), 2018, pp. 4854--4863.

\bibitem{FlowTrack}
Z.~Zhu, W.~Wu, W.~Zou, J.~Yan, End-to-end flow correlation tracking with
  spatial-temporal attention, in: IEEE Conference on Computer Vision and
  Pattern Recognition (CVPR), 2018, pp. 548--557.

\bibitem{SiamRPN}
B.~Li, J.~Yan, W.~Wu, Z.~Zhu, X.~Hu, High performance visual tracking with
  siamese region proposal network, in: IEEE Conference on Computer Vision and
  Pattern Recognition (CVPR), 2018, pp. 8971--8980.

\bibitem{DaSiamRPN}
Z.~Zhu, Q.~Wang, B.~Li, et~al., Distractor-aware siamese networks for visual
  object tracking, in: European Conference on Computer Vision (ECCV), 2018, pp.
  103--119.

\bibitem{VOT2018}
M.~Kristan, A.~Leonardis, J.~Matas, M.~Felsberg, R.~Pflugfelder, et~al., The
  sixth visual object tracking vot2018 challenge results, in: European
  Conference on Computer Vision Workshops (ECCVW), 2018.

\bibitem{HDT}
Y.~Qi, S.~Zhang, L.~Qin, et~al., Hedged deep tracking, in: IEEE Conference on
  Computer Vision and Pattern Recognition (CVPR), 2016, pp. 4303--4311.

\bibitem{STCT}
L.~Wang, W.~Ouyang, X.~Wang, H.~Lu, Stct: Sequentially training convolutional
  networks for visual tracking, in: IEEE Conference on Computer Vision and
  Pattern Recognition (CVPR), 2016, pp. 1373--1381.

\bibitem{Branchout}
B.~Han, J.~Sim, H.~Adam, Branchout: Regularization for online ensemble tracking
  with convolutional neural networks, in: IEEE International Conference on
  Computer Vision (ICCV), 2017, pp. 2217--2224.

\bibitem{TCNN}
H.~Nam, M.~Baek, B.~Han, Modeling and propagating cnns in a tree structure for
  visual tracking, arXiv preprint arXiv:1608.07242.

\bibitem{EDCF}
Q.~Wang, M.~Zhang, et~al., Do not lose the details: reinforced representation
  learning for high performance visual tracking, in: International Joint
  Conferences on Artificial Intelligence (IJCAI), 2018.

\bibitem{SyntheticTIR}
L.~Zhang, A.~Gonzalez-Garcia, et~al., Synthetic data generation for end-to-end
  thermal infrared tracking, Transition on Image Processing (TIP) 28~(4) (2019)
  1837--1850.

\bibitem{pix2pix}
P.~Isola, J.-Y. Zhu, T.~Zhou, A.~A. Efros, Image-to-image translation with
  conditional adversarial networks, in: IEEE Conference on Computer Vision and
  Pattern Recognition (CVPR), 2017, pp. 1125--1134.

\bibitem{RGB-T}
C.~Li, X.~Liang, Y.~Lu, et~al., Rgb-t object tracking: Benchmark and baseline,
  arXiv preprint arXiv:1805.08982.

\bibitem{PTB-TIR}
Q.~Liu, Z.~He, Ptb-tir: A thermal infrared pedestrian tracking benchmark, arXiv
  preprint arXiv:1801.05944.

\bibitem{OSU}
J.~W. Davis, V.~Sharma, Background-subtraction using contour-based fusion of
  thermal and visible imagery, Computer Vision and Image Understanding 106~(2)
  (2007) 162--182.

\bibitem{OTCBVS}
R.~Miezianko, Ieee otcbvs ws series benchmark,
  \url{http://vcipl-okstate.org/pbvs/bench/}, accessed March 4, 2018.

\bibitem{PDT-ATV}
J.~Portmann, S.~Lynen, M.~Chli, R.~Siegwart, People detection and tracking from
  aerial thermal views, in: IEEE International Conference on Robotics and
  Automation (ICRA), 2014, pp. 1794--1800.

\bibitem{Youtube}
Youtube, Youtube thermal videos, \url{www.youtube.com}, accessed March 4, 2018.

\bibitem{SENet}
J.~Hu, L.~Shen, G.~Sun, Squeeze-and-excitation networks, in: IEEE Conference on
  Computer Vision and Pattern Recognition (CVPR), 2018, pp. 7132--7141.

\bibitem{ILSVRC}
O.~Russakovsky, J.~Deng, H.~Su, et~al., Imagenet large scale visual recognition
  challenge, Internation Journal Conmputer Vision (IJCV) 115~(3) (2015)
  211--252.

\bibitem{AlexNet}
A.~Krizhevsky, I.~Sutskever, G.~E. Hinton, Imagenet classification with deep
  convolutional neural networks, in: Advances in neural information processing
  systems, 2012, pp. 1097--1105.

\bibitem{matconvnet}
A.~Vedaldi, K.~Lenc, Matconvnet: Convolutional neural networks for matlab, in:
  ACM International Conference on Multimedia, 2015, pp. 689--692.

\bibitem{ARplot}
M.~Kristan, J.~Matas, A.~Leonardis, et~al., A novel performance evaluation
  methodology for single-target trackers, IEEE Transactions on Pattern Analysis
  and Machine Intelligence 38~(11) (2016) 2137--2155.

\bibitem{VOTperformance1}
L.~{\v{C}}ehovin, M.~Kristan, A.~Leonardis, Is my new tracker really better
  than yours?, in: IEEE Winter Conference on Applications of Computer Vision,
  2014, pp. 540--547.

\bibitem{VOT2015}
M.~Kristan, J.~Matas, A.~Leonardis, M.~Felsberg, et~al., The visual object
  tracking vot2015 challenge results, in: IEEE International Conference on
  Computer Vision Workshops (ICCVW), 2015, pp. 1--23.

\bibitem{DeepMKCF}
M.~Tang, J.~Feng, Multi-kernel correlation filter for visual tracking, in: IEEE
  International Conference on Computer Vision (ICCV), 2015, pp. 3038--3046.

\bibitem{VOT-TIR2016}
M.~Felsberg, M.~Kristan, et~al., The thermal infrared visual object tracking
  vot-tir2016 challenge results, in: European Conference on Computer Vision
  Workshops (ECCVW), 2016, pp. 824--849.

\bibitem{CREST}
Y.~Song, C.~Ma, L.~Gong, et~al., Crest: Convolutional residual learning for
  visual tracking, in: IEEE International Conference on Computer Vision (ICCV),
  2017, pp. 2574--2583.

\bibitem{DeepSTRCF}
F.~Li, C.~Tian, W.~Zuo, L.~Zhang, M.-H. Yang, Learning spatial-temporal
  regularized correlation filters for visual tracking, in: IEEE Conference on
  Computer Vision and Pattern Recognition (CVPR), 2018, pp. 4904--4913.

\bibitem{ECO}
M.~Danelljan, G.~Bhat, F.~Shahbaz~Khan, M.~Felsberg, Eco: efficient convolution
  operators for tracking, in: IEEE Conference on Computer Vision and Pattern
  Recognition (CVPR), 2017, pp. 6638--6646.

\bibitem{VITAL}
Y.~Song, C.~Ma, X.~Wu, L.~Gong, L.~Bao, et~al., Vital: Visual tracking via
  adversarial learning, in: IEEE Conference on Computer Vision and Pattern
  Recognition (CVPR), 2018, pp. 8990--8999.

\end{thebibliography}

\end{document}